\documentclass[a4paper,conference]{IEEEtran}
\usepackage{lipsum}
\usepackage{graphicx}
\usepackage[center]{caption}
\usepackage{subcaption}
\usepackage{amsmath}
\usepackage{amsfonts}
\usepackage{amssymb}
\usepackage{comment}
\usepackage{bm}
\usepackage{xcolor}
\usepackage{float}
\usepackage{adjustbox}
\usepackage{textcomp}

\usepackage[citestyle=numeric-comp,sorting=none,url=false]{biblatex} 
\usepackage{cleveref}

\usepackage{tikz}
\usetikzlibrary{shapes.misc}
\usetikzlibrary{positioning}
\usetikzlibrary{arrows.meta}
\usetikzlibrary{shapes.multipart}
\usetikzlibrary{patterns}

\newcommand{\R}{\mathbb{R}}

\newcommand{\e}{\mathrm{e}}

\graphicspath{{./figs/}} 

\title{Distributed Planning for Rigid Robot Formations using Consensus on the Transformation of a Base Configuration}

\author{Jeppe Heini Mikkelsen, and Matteo Fumagalli \\ Technical University of Denmark, Automation and Control Group}
\date{July 2022}

\begin{document}

\flushbottom

\maketitle
\thispagestyle{plain}
\pagestyle{plain}

\begin{abstract}
    This paper presents a novel planning method that achieves navigation of multi-robot formations in cluttered environments, while maintaining the formation throughout the robots motion. The method utilises a decentralised approach to find feasible formation parameters that guarantees formation constraints for rigid formations. The method proves to be computationally efficient, making it relevant for reactive planning and control of multi-robot systems formation. The method has been tested in a simulation environment to prove feasibility and run-time efficiency.  
\end{abstract}
\section{Introduction}\label{sec:introduction}

Multi-robot operations such as remote swarm operation by a single user, cooperative object transportation, or large-scale area surveying may require a multi-robot system to move in formation. When controlling formations, the robots should move in a reactive manner both to avoid obstacles while keeping formation and for ensuring communication between neighboring robots, while at the same time avoiding self-collisions. This paper presents a multi-robot formation planner that achieves formation control, collision avoidance and communication among the agents in real-time.

Formation planning and control approaches can be divided into centralised and decentralised approaches \cite{Ouyang2023FormationReview}. Centralised approaches compute the robot motions at a central location and subsequently transmit control references to the robots, while decentralised approaches allow all robots in the computation of the formation motion, thus making decentralised approaches generally more robust than centralised methods, due to them not having a single point of failure. Furthermore, robot formations can be either rigid or non-rigid. In rigid formations the robots move in a fixed geometric shape, while in non-rigid formations the formation is permitted to deform. Rigid formations can be less prone to breaking, due to them relying on driving the robots towards an a priori specified formation where relative distances can be guaranteed, but are less flexible in where they can navigate compared to non-rigid formations. Lastly, computational methods for formation planning and control can be optimal or feasible. Optimal methods aim at finding the motion of the robots ensuring that a cost function is minimised, such as distance, time, energy, etc. Feasible methods are used to find a set of motions that is only feasible for the robots to perform, thereby being substantially faster than optimal methods, making them more applicable for real multi-robot systems. 

An early form of distributed feasible non-rigid formation motion control was proposed by Reynolds in 1984 \cite{Reynolds1987FLOCKSMODEL.}, where agents move in formation using local interaction rules. A common approach for this is to use artificial potential fields (APF) \cite{Khatib1985Real-timeRobots}, such as in \cite{Zhang2010DynamicControl,Liu2016FormationSpace,Fan2005Multi-robotNetworks,Zhai2013FormationFramework,Gennaro2006FormationFunctions}. However, APF methods are prone to local minima and can thereby break formation. In \cite{Kanjanawanishkul2010DistributedFormation} the authors propose a distributed feasible rigid formation planning method where the rotation and translation of a base configuration is found by calculating the first principal component of the robot positions using a consensus algorithm, and an assignment of robots to the formation is found using a distributed negotiation algorithm. In \cite{Montijano2014EfficientOptimization} a similar, but optimal, approach for finding the formation and assignment that minimises the distance travelled by the robots to reach the formation was proposed. In \cite{Alonso-Mora2016DistributedConsensus} the authors find the optimal assignment, scaling, translation, and rotation for navigating a rigid formation of drones towards a goal in a cluttered environment, in a distributed manner, while combining individual robot collision avoidance and local planning algorithms to navigate the robots to the desired formation. However, the method does not ensure that robots remain in formation while in transit from one formation to another. 

In this paper we present a novel distributed, feasible planner for rigid formations. Similarly to \cite{Kanjanawanishkul2010DistributedFormation,Montijano2014EfficientOptimization,Alonso-Mora2016DistributedConsensus,Alonso-Mora2015Multi-robotProgramming}, our method finds the scaling, rotation, and translation of a base configuration. However, we propose a continuous, fast, light-weight, distributed planner that allows robots moving in a rigid formation while obeying constraints on the formation parameters. We achieve this by mapping the desired velocities of the robots into the parameter space of a formation transformation, performing consensus and constraint steps, and mapping back to the velocity space of the robots. Furthermore, our method continuously keeps the robots 
in formation. 

In this paper the following notation style is used: Lowercase italic symbols $x$ are variables, bold lowercase italic symbols $\bm{x}$ are vectors, bold uppercase symbols $\mathbf{X}$ are matrices, and calligraphic symbols $\mathcal{X}$ are sets. $\R_{\geq0}$ and $\R_{>0}$ denotes positive and strictly positive real numbers respectively.

\section{Problem Description and Approach}\label{sec:problem}
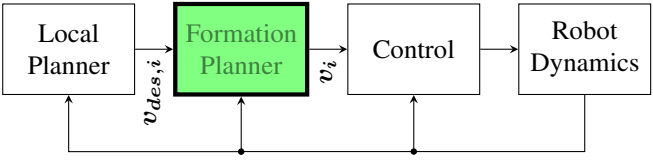
\begin{figure}[t!]
    \centering
    \begin{tikzpicture}
        \node [draw,
            fill=white,
            minimum width=1.5cm,
            minimum height=1.2cm,
            text width=1.5cm,
            align=center
        ]  (local_planner) at (0,0) {Local Planner};

        \node [draw,
            fill=green,
            fill opacity=0.5,
            text opacity=1,
            minimum width=1.5cm,
            minimum height=1.2cm,
            text width=1.5cm,
            align=center,
            right = 0.5cm of local_planner,
            ultra thick
        ] (formation_planner) {Formation Planner};

        \node [draw,
            fill=white,
            minimum width=1.5cm,
            minimum height=1.2cm,
            text width=1.5cm,
            align=center,
            right = 0.5cm of formation_planner
        ] (control) {Control};

        \node [draw,
            fill=white,
            minimum width=1.5cm,
            minimum height=1.2cm,
            text width=1.5cm,
            align=center,
            right = 0.5cm of control
        ] (robot) {Robot Dynamics};

        \draw[-stealth] (local_planner.east) -- (formation_planner.west) node[midway,rotate=90,left]{$\bm{v_{des,i}}$};

        \draw[-stealth] (formation_planner.east) -- (control.west) node[midway,rotate=90,left]{$\bm{v_{i}}$};

        \draw[-stealth] (control.east) -- (robot.west);

        \draw[-stealth] (robot.south) -- ++ (0,-0.75) -| (control.south);

        \draw[-stealth] (control.south) ++ (0,-0.75) -| (formation_planner.south);

        \draw[-stealth] (formation_planner.south) ++ (0,-0.73) -| (local_planner.south);

        \filldraw[black] (control.south) ++ (0,-0.75) circle (1pt);

        \filldraw[black] (formation_planner.south) ++ (0,-0.73) circle (1pt);
        
    \end{tikzpicture}
    \caption{System architecture: Each robot has a local planner, a formation planner and a control system.}
    \label{fig:sys_architecture}
\end{figure}

Consider a swarm of $N$ planar robots, $\mathcal{V} = \{1,\dots,N\}$, where the position of robot $i$ is denoted by $\bm{p_i} \in \R^2$. It is assumed that the robots are holonomic and that each robot has a local kinematic controller, ensuring that it is able to track a reference velocity. Therefore, the dynamics of each robot is represented using the following single-integrator model:
\begin{equation}
    \frac{d}{dt}\bm{p_i} = \bm{v_i}, \quad \forall i \in \mathcal{V}.
\end{equation}
It is assumed that the robots exchange information with each other through wireless communication. The communication network can be modelled as an undirected dynamic graph $\mathcal{G}(t) = (\mathcal{V},\mathcal{E}(t))$, where $\mathcal{E}(t) = \{(i,j)\in\mathcal{V}\times\mathcal{V} \ | \ i \neq j \ \wedge \ ||\bm{p_i} - \bm{p_j}||_2 \leq r_c\}$ are time-varying communication links between robot pairs, with $r_c$ being the communication range. $\mathcal{N}_i(t) = \{j \in \mathcal{V} \ | \ (i,j) \in \mathcal{E}(t)\}$ is the neighbour set of robot $i$, i.e., the robots with which robot $i$ has a direct communication link. Furthermore, there is a distance $r_d \leq r_c$ wherein communication performance is assumed to be perfect or near perfect, and after which it starts to degrade. The goal of this paper is to derive a distributed algorithm that finds the velocities $\bm{v_i}$ for each robot in the swarm, such that they each attempt to track a local desired velocity $\bm{v_{des,i}}$ while remaining in formation, maintaining communication and avoiding collisions. This is achieved by injecting an intermediary formation planner between the local planner and the controller on each robot, where the desired velocities are supplied by the local planners; see \cref{fig:sys_architecture}. The local planners could be a variety of planners and is not within the scope of this paper.
\section{Formation Transformation}\label{sec:transformation}
The positions of the robots in the formation is parameterised through a transformation of a base configuration. Consider a base configuration $\mathcal{B}=\{\bm{c_1},\dots,\bm{c_N}\}$, where $\bm{c_i}\in\R^2$ is the position in the base configuration associated with robot $i$. The position of robot $i$ in the formation is then found according to the following scaling, rotation, and translation.
\begin{equation}\label{eq:transformation}
    \bm{p_i} = \mathbf{R}\mathbf{S}\bm{c_i} + \bm{t},
\end{equation}
where $\mathbf{S}\in\R_{>0}^{2\times2}$ is a strictly positive diagonal scaling matrix, $\mathbf{R}\in SO(2)$ is a rotation matrix, and $\bm{t}\in\R^2$ is a translation vector; see \cref{fig:transformation}
\begin{equation}\label{eq:transformation_components}
    \mathbf{R} = 
    \begin{bmatrix}
        \cos\varphi & -\sin\varphi \\
        \sin\varphi &  \cos\varphi
    \end{bmatrix}, \
    \mathbf{S} =
    \begin{bmatrix}
        s_x & 0 \\
        0   & s_y
    \end{bmatrix}, \
    \bm{t} =
    \begin{bmatrix}
        t_x \\
        t_y
    \end{bmatrix}.
\end{equation}
The parameter vector of the transformation is denoted as
\begin{equation}
    \begin{gathered}
    \bm{\eta} = (\varphi,\bm{s},\bm{t}) \in \R^5, \\ \quad \bm{s} = (s_x,s_y)\in\R^2, \quad \bm{t} = (t_x,t_y)\in\R^2.
    \end{gathered}
\end{equation}
\tikzset{cross/.style={cross out, draw=black, fill=none, minimum size=2*(#1-\pgflinewidth), inner sep=0pt, outer sep=0pt}, cross/.default={5pt}}
\begin{figure}[t!]
    \centering
    \begin{tikzpicture}[scale=0.75]

        \draw[thick,-stealth] (-2,0) -- (6,0) node[anchor=north east]{$x$};
        \draw[thick,-stealth] (0,-2) -- (0,3) node[anchor=north east]{$y$};
        \filldraw[black] (0,0) circle (1pt);        
        \draw[ultra thick] (-1,-1) circle (4pt);
        \draw[ultra thick] (-1,0) circle (4pt);
        \draw[ultra thick] (-1,1) circle (4pt);
        \draw[ultra thick] (0,-1) circle (4pt);
        \draw[ultra thick] (0,0) circle (4pt);
        \draw[ultra thick] (0,1) circle (4pt);
        \draw[ultra thick] (1,-1) circle (4pt);
        \draw[ultra thick] (1,0) circle (4pt);
        \draw[ultra thick] (1,1) circle (4pt);
        
        \draw[thick,dashed] (1,3) -- (5,-1);
        \draw[thick,dashed] (4.5,2.5) -- (1.5,-0.5);
        \draw[thick,dashed] (3,1) -- (5,1);


        \draw[ultra thick,red] (3.7071,-1.1213) circle (4pt);
        \draw[ultra thick,red] (2.2929,0.2929) circle (4pt);
        \draw[ultra thick,red] (0.8787,1.7071) circle (4pt);
        \draw[ultra thick,red] (4.4142,-0.4142) circle (4pt);
        \draw[ultra thick,red] (3.0000,1.0000) circle (4pt);
        \draw[ultra thick,red] (1.5858,2.4142) circle (4pt);
        \draw[ultra thick,red] (5.1213,0.2929) circle (4pt);
        \draw[ultra thick,red] (3.7071,1.7071) circle (4pt);
        \draw[ultra thick,red] (2.2929,3.1213) circle (4pt);

        \draw[thick,->] (5,1) arc (0:45:2);
        \node at (5.15,1.9) {$\varphi$};
        \draw[thick,->] (0,0) -- (3,1);
        \node at (2.5,1.1) {$\bm{t}$};
        \draw[thick,->] (3,1) -- (3.71,1.71);
        \node at (3.21,1.71) {$s_x$};
        \draw[thick,->] (3,1) -- (1.59,2.41);
        \node at (2.09,2.41) {$s_y$};

    \end{tikzpicture}
    \caption{Transformation of a unit grid swarm configuration with $\varphi=\pi/4$, $s_x=1$, $s_y=2$, $t_x=3$, and $t_y=1$. \textbf{Black O}: base configuration. \textcolor{red}{\textbf{Red O}}: transformed base configuration.}
    \label{fig:transformation}
\end{figure}
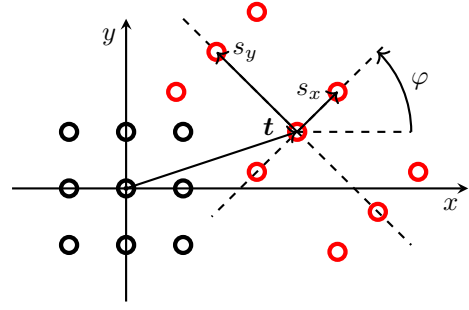
The base configuration $\mathcal{B}$ is determined a priori and can have any desired shape, e.g., grid, triangular, hexagonal, etc., see \cref{fig:base_configurations}.
\begin{figure}[t!]
    \centering
    \begin{subfigure}[b]{0.3\linewidth}
        \centering
        \begin{tikzpicture}
            \draw[thick,-stealth] (-1,0) -- (1,0) node[anchor=north east]{$x$};
            \draw[thick,-stealth] (0,-1) -- (0,1) node[anchor=north east]{$y$};
            \filldraw[black] (0,0) circle (1pt);  
            \draw[ultra thick,black] (0,0) circle (3pt);
            \draw[ultra thick,black] (0.5,0) circle (3pt);
            \draw[ultra thick,black] (-0.5,0) circle (3pt);
            \draw[ultra thick,black] (0,0.5) circle (3pt);
            \draw[ultra thick,black] (0,-0.5) circle (3pt);
            \draw[ultra thick,black] (0.5,0.5) circle (3pt);
            \draw[ultra thick,black] (0.5,-0.5) circle (3pt);
            \draw[ultra thick,black] (-0.5,0.5) circle (3pt);
            \draw[ultra thick,black] (-0.5,-0.5) circle (3pt);
        \end{tikzpicture}
        \caption{grid}
        \label{fig:grid}
    \end{subfigure}
    \begin{subfigure}[b]{0.3\linewidth}
        \centering
        \begin{tikzpicture}
            \draw[thick,-stealth] (-1,0) -- (1,0) node[anchor=north east]{$x$};
            \draw[thick,-stealth] (0,-1) -- (0,1) node[anchor=north east]{$y$};
            \filldraw[black] (0,0) circle (1pt);  
            \draw[ultra thick,black] (0,0.5) circle (3pt);
            \draw[ultra thick,black] (-0.25,0) circle (3pt);
            \draw[ultra thick,black] (0.25,0) circle (3pt);
            \draw[ultra thick,black] (0,-0.5) circle (3pt);
            \draw[ultra thick,black] (-0.5,-0.5) circle (3pt);
            \draw[ultra thick,black] (0.5,-0.5) circle (3pt);
        \end{tikzpicture}
        \caption{triangular}
        \label{fig:triangular}
    \end{subfigure}
    \begin{subfigure}[b]{0.3\linewidth}
        \centering
        \begin{tikzpicture}
            \draw[thick,-stealth] (-1,0) -- (1,0) node[anchor=north east]{$x$};
            \draw[thick,-stealth] (0,-1) -- (0,1) node[anchor=north east]{$y$};
            \filldraw[black] (0,0) circle (1pt);  
            \draw[ultra thick,black] (0,0.5) circle (3pt);
            \draw[ultra thick,black] (-0.433,0.25) circle (3pt);
            \draw[ultra thick,black] (-0.433,-0.25) circle (3pt);
            \draw[ultra thick,black] (0,-0.5) circle (3pt);
            \draw[ultra thick,black] (0.433,0.25) circle (3pt);
            \draw[ultra thick,black] (0.433,-0.25) circle (3pt);
        \end{tikzpicture}
        \caption{hexagonal}
        \label{fig:hexagonal}
    \end{subfigure}
        \caption{Examples of three different base configurations.}
        \label{fig:base_configurations}
\end{figure}
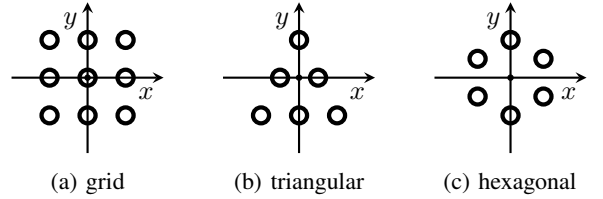

\section{Method}\label{sec:method}
Having found a transformation that expresses the position of each robot in the formation, the motion that ensures that each robot stays in formation can be found through a four steps approach: \textbf{tracking, consensus, constraint satisfaction and recovering velocity}.
To ensure that the robots find their velocities in a distributed way, each robot $i$ carries an instance of the transformation parameters, denoted as $\bm{\eta_i}$, and performs the steps locally.
\subsection*{Step 1: Tracking}
In the first step, the time derivative of the parameters, which ensures that each robot tracks its desired velocity, is found. Using the chain rule, the time derivative of the position of robot $i$ in the swarm, with respect to the time derivative of its parameters, can be expressed as
\begin{equation}
    \frac{d}{dt}\bm{p_i} = \mathbf{J_{\bm{\eta_i}}}\frac{d}{dt}\bm{\eta_i},
\end{equation}
where $\mathbf{J_{\bm{\eta_i}}}$ is the Jacobian of the transformation in \eqref{eq:transformation} with respect to $\bm{\eta_i}$,
\begin{multline}\label{eq:jacobian}
    \mathbf{J}_{\bm{\eta_i}} = \left[
    \begin{matrix}
        -\sin\varphi_i s_{x,i} c_{x,i} - \cos\varphi_i s_{y,i} c_{y,i} \\
        \cos\varphi_i s_{x,i} c_{x,i} -\sin\varphi_i s_{y,i} c_{y,i}
    \end{matrix} \right. \dots\\
    \left.
    \begin{matrix}
        \cos\varphi_i c_{x,i} & -\sin\varphi_i c_{y,i} & 1 & 0 \\
        \sin\varphi_i c_{x,i} & \cos\varphi_i c_{y,i} & 0 & 1
    \end{matrix} \right].
\end{multline}
From this, the time derivative of the parameters for robot $i$ can be found as
\begin{equation}\label{eq:step1}
    \frac{d}{dt}\bm{\eta_i} = \mathbf{J}^+_{\bm{\eta_i}}\bm{v_{des,i}},
\end{equation}
where $(\cdot)^+$ denotes the right Moore-Penrose pseudo-inverse, and $\bm{v_{des,i}}$ is the desired velocity of robot $i$. Since the rows of the Jacobian are linearly independent, the pseudo-inverse can be computed as
\begin{equation}
    \mathbf{J_{\bm{\eta_i}}^+} = \mathbf{J_{\bm{\eta_i}}^\top}(\mathbf{J_{\bm{\eta_i}}}\mathbf{J_{\bm{\eta_i}}}^\top)^{-1}.
\end{equation}
However, since the Jacobian is underdetermined and since the individual desired velocities may not conform to a feasible formation motion, applying the update in \eqref{eq:step1} does not result in a unique solution across the robots and therefore there will be discrepancies among the robots as to the parameters.
\subsection*{Step 2: Consensus}
To ensure that the robots find the same solution, a consensus step is applied to \eqref{eq:step1}
\begin{equation}
    \frac{d}{dt}\bm{\eta_i} = \mathbf{J}^+_{\bm{\eta_i}}\bm{v_{des,i}} \underbrace{-\lambda_i\sum_{j\in\mathcal{N}_i} (\bm{\eta_i} - \bm{\eta_j})}_{\textbf{consensus step}}, \quad \lambda_i \in \R_{>0},
\end{equation}
where $\lambda_i$ is a strictly positive multiplier that determines how fast consensus is reached. Applying this step drives the solutions of the robots together, ensuring that they remain in formation.
\subsection*{Step 3: Constraint Satisfaction}
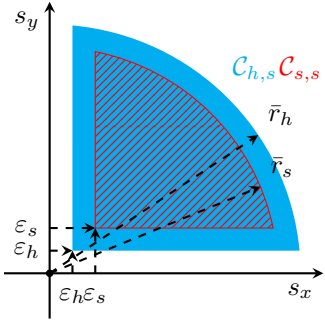
\begin{figure}[t!]
    \centering
    \begin{tikzpicture}[scale=1.2]
        \draw[thick,-stealth] (-0.5,0) -- (3,0) node[anchor=north east]{$s_x$};
        \draw[thick,-stealth] (0,-0.5) -- (0,3) node[anchor=north east]{$s_y$};
        \filldraw[black] (0,0) circle (1pt);

        \fill[cyan] (0.25,0.25) -- (2.7386,0.25) arc (5.2159:84.7841:2.75) -- cycle;
        \def\mypath{(0.5,0.5) -- (2.4495,0.5) arc (11.5370:78.4630:2.5) -- cycle}; 
        \draw[red] \mypath;
        \pattern[pattern color=red,pattern=north east lines] \mypath;

        \node[] at (-0.25,0.5) {$\varepsilon_s$};
        \draw[-stealth,thick,dashed] (0,0.5) -- (0.5,0.5);
        \node[] at (0.5,-0.25) {$\varepsilon_s$};
        \draw[-stealth,thick,dashed] (0.5,0) -- (0.5,0.5);

        \node[] at (-0.25,0.25) {$\varepsilon_h$};
        \draw[-stealth,thick,dashed] (0,0.25) -- (0.25,0.25);
        \node[] at (0.25,-0.25) {$\varepsilon_h$};
        \draw[-stealth,thick,dashed] (0.25,0) -- (0.25,0.25);

        \node[cyan] at (2.25,2.25) {$\mathcal{C}_{h,s}$};
        \node[red] at (2.75,2.25) {$\mathcal{C}_{s,s}$};

        \draw[thick,-stealth,dashed] (0,0) -- (2.3097,0.9567) node[anchor=south west] {$\bar{r}_s$};

        \draw[thick,-stealth,dashed] (0,0) -- (2.2865,1.5278) node[anchor=south west] {$\bar{r}_h$};

    \end{tikzpicture}
    \caption{Soft and hard constraint set on parameter $s$.}
    \label{fig:constraint_sets}
\end{figure}
The two prior steps allow unbounded scaling, which can result in loss of communication or collisions. To remedy this, an additional step is applied to constrain the solution. Due to the communication model, there is a range wherein the robots are assumed to have perfect, or near perfect, communication. After that, the communication attenuates until it ceases to work. Furthermore, there is a minimum distance that robots prefer to have to each other and a minimum distance that they have to keep from each other to avoid collisions. Therefore, a soft constraint and a hard constraint on the transformation parameters are introduced. The set in which the hard constraint requires the parameters to be within is denoted as $\mathcal{C}_{h} \in \R^5$, and the set in which the soft constraint prefers the parameters to be within is denoted as $\mathcal{C}_{s} \in \R^5$, with $\mathcal{C}_{s} \subseteq \mathcal{C}_{h}$.
Since it is only the scaling parameter that has an influence on the relative distances between the robots, as $\mathbf{R},\bm{t} \in SE(2)$, the rotation and translation parameters, $\varphi$ and $\bm{t}$, are unconstrained, i.e., $\mathcal{C}_{s,\phi},\mathcal{C}_{h,\phi} \in \R$ and $\mathcal{C}_{s,t},\mathcal{C}_{h,t} \in \R^2$. The soft and hard constraint sets on the scaling parameter $\bm{s}$, $\mathcal{C}_{s,s},\mathcal{C}_{h,s}\subseteq\R^2_{>0}$, both consist of a subset of a quarter circle in the positive quadrant; see \cref{fig:constraint_sets}. To help avoid collisions, it is preferred that the scaling in $x$ and $y$ is greater than the lower bound $\varepsilon_s\in\R_{>0}$ and it is required to be greater than the lower bound $\varepsilon_h\in\R_{>0}$, where $\varepsilon_s \geq \varepsilon_h$. To help ensure communication, the Euclidean norm of the scaling is preferred to be smaller than an upper bound $\bar{r}_s\in\R_{>0}$ and is required to be smaller than an upper bound $\bar{r}_h\in\R_{>0}$, where $\bar{r}_s \leq \bar{r}_h$.
\subsubsection{Soft Constraint}
The soft constraint step attempts to drive the parameters toward the soft constraint set $\mathcal{C}_{s}$ as
\begin{multline}\label{eq:soft_constraint_satisfaction}
    \frac{d}{dt}\bm{\eta_i} = \mathbf{J}^+_{\bm{\eta_i}}\bm{v_{des,i}} - \lambda_i \sum_{j\in\mathcal{N}_i}(\bm{\eta_i} - \bm{\eta_j})\dots \\ \underbrace{- \mu_i(\bm{\eta_i} - \text{proj}(\bm{\eta_i},\mathcal{C}_s))}_{\textbf{soft constraint step}},
\end{multline}
where $\mu_i \in \R_{\geq0}$ is a positive penalty multiplier that determines how hard the soft constraint attempts to drive the parameters into $\mathcal{C}_{s}$, and $\text{proj}(\bm{\eta_i},\mathcal{C}_s) \in \R^5$ is a projection of the parameters onto $\mathcal{C}_{s}$
\begin{equation}
    \text{proj}(\bm{\eta_i},\mathcal{C}) = (\varphi_{proj},\bm{s_{proj}},\bm{t_{proj}}).
\end{equation}
Since the rotation and translation is unconstrained, their projection is set to their current value,
\begin{equation}
    \varphi_{proj} = \varphi, \quad \bm{t_{proj}} = \bm{t}.  
\end{equation}
\begin{figure}[t!]
    \centering
    \begin{tikzpicture}[scale=1.2]
        \fill[red] (0.5,0.5) -- (2.4495,0.5) arc (11.5370:78.4630:2.5) -- cycle; 
        \fill[pink] (-0.5,-0.5) rectangle (0.5,0.5);
        \fill[yellow] (0.5,-0.5) rectangle (2.4495,0.5);
        \fill[cyan] (-0.5,0.5) rectangle (0.5,2.4495);
        \fill[magenta] (2.4495,-0.5) rectangle (3,0.5);
        \fill[orange] (-0.5,2.4495) rectangle (0.5,3);
        \fill[lime] (3,0.5) -- (2.4495,0.5) arc (11.5370:78.4630:2.5) -- (0.5,3) -- (3,3) -- cycle;

        \filldraw[black] (0.25,0.25) circle (1pt) node[anchor=west]{\footnotesize\textit{1}};
        \draw[thick,-stealth] (0.25,0.25) -- (0.5,0.5);

        \filldraw[black] (1.5,0.25) circle (1pt) node[anchor=west]{\footnotesize\textit{2}};
        \draw[thick,-stealth] (1.5,0.25) -- (1.5,0.5);

        \filldraw[black] (0.25,1.5) circle (1pt) node[anchor=north east]{\footnotesize\textit{3}};
        \draw[thick,-stealth] (0.25,1.5) -- (0.5,1.5);
        
        \filldraw[black] (2.75,0.25) circle (1pt) node[anchor=west]{\footnotesize\textit{4}};
        \draw[thick,-stealth] (2.75,0.25) -- (2.4495,0.5);
        
        \filldraw[black] (0.25,2.75) circle (1pt) node[anchor=north east]{\footnotesize\textit{5}};
        \draw[thick,-stealth] (0.25,2.75) -- (0.5,2.4495);

        \filldraw[black] (2.25,2.25) circle (1pt) node[anchor=north west]{\footnotesize\textit{6}};
        \draw[thick,-stealth] (2.25,2.25) -- (1.768,1.768);
        
        \filldraw[black] (1.25,1.25) circle (1pt) node[anchor=north west]{\footnotesize\textit{7}};

        \node[] at (1,1) {$\mathcal{C}_{s,s}$};

        \draw[thick,-stealth] (-0.5,0) -- (3,0) node[anchor=north east]{$s_x$};
        \draw[thick,-stealth] (0,-0.5) -- (0,3) node[anchor=north east]{$s_y$};
        \filldraw[black] (0,0) circle (1pt);
        
        \draw[thick,-stealth,dashed] (0,0) -- (1.25,2.165) node[anchor=south west] {$\bar{r}_s$};
        
        \node[] at (-0.25,0.5) {$\varepsilon_s$};
        \node[] at (-0.25,2.4495) {$\delta_s$};
        \node[] at (0.5,-0.25) {$\varepsilon_s$};
        \node[] at (2.4495,-0.25) {$\delta_s$};

    \end{tikzpicture}
    \caption{Set of preferable scalings $\mathcal{C}_{s,s}$ and the projections 1-7 for the seven cases in \eqref{eq:projection_cases} in order of appearance.}
    \label{fig:constraint_satisfaction_set}
\end{figure}
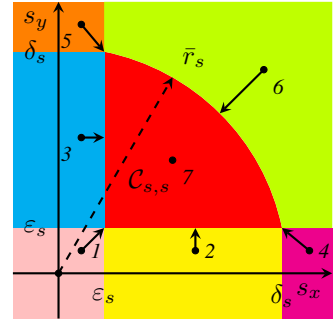
\begin{figure}[t!]
    \centering
    \begin{tikzpicture}[scale=1.2]
        \draw[thick,-stealth] (-0.5,0) -- (3,0) node[anchor=north east]{$s_x$};
        \draw[thick,-stealth] (0,-0.5) -- (0,3) node[anchor=north east]{$s_y$};
        \filldraw[black] (0,0) circle (1pt);

        \fill[cyan] (0.25,0.25) -- (2.7386,0.25) arc (5.2159:84.7841:2.75) -- cycle;

        \node[] at (-0.25,0.25) {$\varepsilon_h$};
        \draw[-stealth,thick,dashed] (0,0.25) -- (0.25,0.25);
        \node[] at (0.25,-0.25) {$\varepsilon_h$};
        \draw[-stealth,thick,dashed] (0.25,0) -- (0.25,0.25);

        \node at (0.75,1.25) {$\mathcal{C}_{h,s}$};

        \draw[-stealth,thick] (1.3,1.7) -- (1.7,2.5) node[anchor=west] {$\frac{d}{dt}\bm{s}$};
        \draw[red,-stealth,thick] (1.3,1.7) -- (1.5766,2.2532) node[red,anchor=east] {$a_s\frac{d}{dt}\bm{s}$};

        \filldraw[black] (1.3,1.7) circle (1pt) node[anchor=north west] {$\bm{s}$};

        \draw[thick,-stealth,dashed] (0,0) -- (2.2865,1.5278) node[anchor=south west] {$\bar{r}_{h}$};

    \end{tikzpicture}
    \caption{}
    \label{fig:hard_constraint_set}
\end{figure}
There are seven different cases for how the projection of the scaling parameter $\bm{s}$ is performed,
\begin{equation}\label{eq:projection_cases}
    \bm{s_{proj}} = \begin{cases}
        (\varepsilon_s,\varepsilon_s)  & \text{if } s_x < \varepsilon_s \wedge s_y < \varepsilon_s, \\
        (s_x,\varepsilon_s)  & \text{if } s_x \geq \varepsilon_s \wedge s_x < \delta_s \wedge s_y < \varepsilon_s, \\
        (\varepsilon_s,s_y)  & \text{if } s_x  < \varepsilon_s \wedge s_y \geq \varepsilon_s \wedge s_y < \delta_s, \\
        (\delta_s,\varepsilon_s)  & \text{if } s_x \geq \delta_s \wedge s_y < \varepsilon_s, \\
        (\varepsilon_s,\delta_s)  & \text{if } s_x < \varepsilon_s \wedge s_y \geq \delta_s, \\
        r_{max,s}\dfrac{\bm{s}}{||\bm{s}||_2} \ & \text{if } s_x \geq \varepsilon_s \wedge s_y \geq \varepsilon_s \wedge ||\bm{s}||_2 > \bar{r}_s, \\
        \bm{s}  & \text{otherwise},
    \end{cases}
\end{equation}
where $\delta_s = \sqrt{\bar{r}_s^2 - \varepsilon_s^2}$, see \cref{fig:constraint_satisfaction_set}.
\subsubsection{Hard Constraint}
Having applied the soft constraint, the hard constraint needs to be applied. The hard constraint is applied by scaling the parameter derivative by a matrix $\mathbf{A_i}$
\begin{multline}\label{eq:hard_constraint}
    \frac{d}{dt}\bm{\eta_i} \leftarrow \mathbf{A_i}\frac{d}{dt}\bm{\eta_i} = \mathbf{A_i}\mathbf{J}^+_{\bm{\eta_i}}\bm{v_{des,i}} \dots \\ - \lambda_i \mathbf{A_i}\sum_{j\in\mathcal{N}_i}(\bm{\eta_i} - \bm{\eta_j}) - \mu_i\mathbf{A_i}(\bm{\eta_i} - \text{proj}(\bm{\eta_i},\mathcal{C}_s))
\end{multline}
where $\mathbf{A_i}$ is a diagonal matrix
\begin{equation}
    \mathbf{A_i} = \text{diag}(a_{\varphi,i},a_{s,i}\mathbf{1}_{2\times1},a_{t,i}\mathbf{1}_{2\times1}).
\end{equation}
Since the rotation and translation are unconstrained, their scaling parameter is one,
\begin{equation}
    a_{\varphi,i} = 1, \quad a_{t,i} = 1.
\end{equation}
The scaling parameter derivative is scaled such that it always lies within the hard constraint set $\mathcal{C}_{h,s}$, as
\begin{equation}\label{eq:hard_constraint_opt_prob}
    \begin{aligned}
        a_{s,i} =& \max \ \alpha_s, \\
        &s.t. \ \bm{s_i} + \alpha_s\frac{d}{dt}\bm{s_i} \in \mathcal{C}_{h,s}, \\
        &\alpha_s \in [0,1].
    \end{aligned} 
\end{equation}
This ensures that the scaling parameters cannot exit the hard constraint set, as, if they try to exit, their derivative converges to zero as they approach the edge of the set; see \cref{fig:hard_constraint_set}. For the specific hard constraint set in \cref{fig:constraint_sets}, \cref{eq:hard_constraint_opt_prob} can be solved as
\begin{equation}
    a_{s,i} = \min_{\geq 0} \left(1,(\varepsilon_h - \bm{s_i})\oslash\frac{d}{dt}\bm{s_i},\frac{- \beta \pm \sqrt{\beta^2 - 4\alpha\gamma}}{2\alpha}\right),
\end{equation}
where $\min_{\geq 0}$ denotes the smallest positive element, $\oslash$ denotes the Hadamard division operator, and
\begin{gather}
    \alpha = \left(\frac{d}{dt}\bm{s_i}^\top\right)\cdot\left(\frac{d}{dt}\bm{s_i}\right),  \\
    \beta = 2\left(\frac{d}{dt}\bm{s_i}^\top\right)\cdot\bm{s_i}, \\ 
    \gamma = \bm{s_i}^\top\cdot\bm{s_i} - \bar{r}_h^2. 
\end{gather}
\subsection*{Step 4: Recovering Velocity}
Since the local controllers on the robots work in the velocity space, the parameter derivative must be transformed back to a velocity. This is achieved by pre-multiplying \eqref{eq:hard_constraint} with the Jacobian in \eqref{eq:jacobian}. However, this does not ensure that the robots return to formation if it is broken due to unforeseen perturbations. Therefore, an additional term is added that drives the robots towards their current desired formation
\begin{equation}\label{eq:recovered_velocity}
    \begin{aligned}
        \bm{v_i} =& \mathbf{J_{\bm{\eta_i}}}\frac{d}{dt}\bm{\eta_i} \underbrace{- K_{i}(\bm{p_i} - (\mathbf{R_i}\mathbf{S_i}\bm{c_i} + \bm{t_i}))}_{\textbf{perturbation rejection}} \\
        =& \mathbf{\Gamma_i}\bm{v_{des,i}} - \lambda_i\mathbf{J_{\bm{\eta_i}} A_i} \sum_{j\in\mathcal{N}_i}(\bm{\eta_i} - \bm{\eta_j}) \dots \\
        &- \mu_i \mathbf{J_{\bm{\eta_i}} A_i}(\bm{\eta_i} - \text{proj}(\bm{\eta_i},\mathcal{C}_s)) \dots \\ &- K_{i}(\bm{p_i} - (\mathbf{R_i}\mathbf{S_i}\bm{c_i} + \bm{t_i})),
    \end{aligned}
\end{equation}
where $\mathbf{\Gamma_i} = \mathbf{J_{\bm{\eta_i}}}\mathbf{A_i}\mathbf{J^+_{\bm{\eta_i}}}$, and $K_{i} \in \R_+$ is a positive feedback gain that determines how fast the robot return to formation. The recovered velocity in \cref{eq:recovered_velocity} consists of the original desired velocity $\bm{v_{des,i}}$ with the corrections ensuring that the robots move in formation, that the formation stays within constraints and that the robots return to formation in case of perturbations. $\lambda_i$ can be interpreted as a stiffness coefficient determining how much the robots keep formation, and $\mu_i$ a stiffness coefficient determining how much the robots want to obey the soft constraint. 
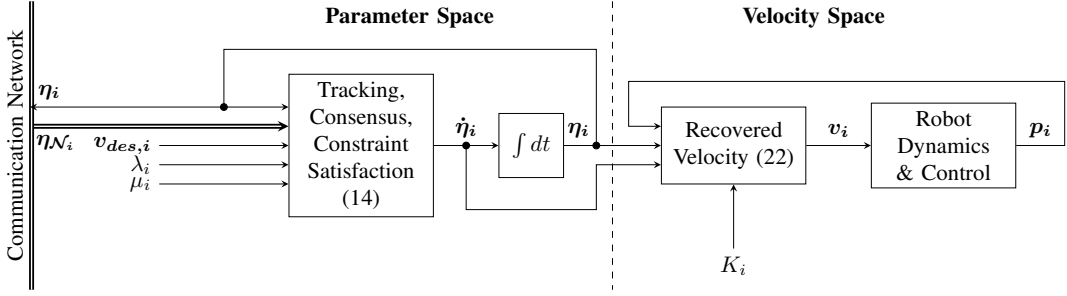
\begin{figure*}[t!]
    \centering
    \scalebox{0.85}{
    \begin{tikzpicture}
        \node [draw,
            fill=white,
            minimum width=2cm,
            minimum height=2cm,
            text width=2 cm,
            align=center
        ]  (parameter_derivative) at (0,0) {Tracking, Consensus, Constraint Satisfaction \eqref{eq:hard_constraint}};

        \node [draw,
            fill=white, 
            minimum width=1cm, 
            minimum height=1cm,
            right = 1cm of parameter_derivative
        ] (integrator_1) {$\int  dt$};

        \draw[-stealth] (parameter_derivative.east) -- (integrator_1.west) node[midway,above]{$\bm{\dot{\eta}_i}$};

        \node [draw,
            fill=white, 
            minimum width=2cm, 
            minimum height=1.2cm,
            text width=2 cm,
            align=center,
            right = 1.5cm of integrator_1
        ] (recovered_velocity) {Recovered Velocity \eqref{eq:recovered_velocity}};

        \node [draw,
            fill=white, 
            minimum width=2cm, 
            minimum height=1.2cm,
            right = 1cm of recovered_velocity,
            text width=2cm,
            align=center
        ] (robot_dynamics) {Robot Dynamics \& Control};

        \draw[-stealth] (integrator_1.east) -- (recovered_velocity.west) node[pos=0.15,above]{$\bm{\eta_i}$};

        \draw[-stealth] (parameter_derivative.east) ++ (0.5,0) -- ++ (0,-1) -- ([shift=({-1,-1})]recovered_velocity.west) |- ([shift=({0,-0.3})]recovered_velocity.west);

        \draw[-stealth] (recovered_velocity.east) -- (robot_dynamics.west) node[midway,above]{$\bm{v_i}$};

        \draw[-stealth] (robot_dynamics.east) -- ++ (0.75,0) node[pos=0.5,above]{$\bm{p_i}$} -- ++ (0,1) -- ([shift=({-0.5,1})]recovered_velocity.west) |- ([shift=({0,0.3})]recovered_velocity.west);

        \draw[-stealth] ([shift=({0.5,0})]integrator_1.east) -- ++(0,1.5) -- ([shift=({-1,1.5})]parameter_derivative.west) |- ([shift=({0,0.6})]parameter_derivative.west);

        \draw[-stealth] ([shift=({-2,0})]parameter_derivative.west) --  (parameter_derivative.west) node[pos=0,left]{$\bm{v_{des,i}}$};

        \draw[-stealth] ([shift=({-2,-0.3})]parameter_derivative.west) --  ([shift=({0,-0.3})]parameter_derivative.west) node[pos=0,left]{$\lambda_i$};

        \draw[-stealth] ([shift=({-2,-0.6})]parameter_derivative.west) --  ([shift=({0,-0.6})]parameter_derivative.west) node[pos=0,left]{$\mu_i$};

        \draw[-stealth] ([shift=({0,-1})]recovered_velocity.south) --  (recovered_velocity.south) node[pos=0,below]{$K_{i}$};

        \node at (0.75,2) {\textbf{Parameter Space}};

        \node at (7,2) {\textbf{Velocity Space}};

        \draw[dashed] ([shift=({-0.75,2.25})]recovered_velocity.west) -- ([shift=({-0.75,-2.25})]recovered_velocity.west);

        \draw[thick,double,-stealth] ([shift=({-4,0.3})]parameter_derivative.west) --  ([shift=({0,0.3})]parameter_derivative.west) node[pos=0.1,below]{$\bm{\eta_{\mathcal{N}_i}}$};

        \draw[thick,double] (-5.1,2.25) -- (-5.1,-2.25) node[midway,above,rotate=90]{Communication Network};

        \draw[-stealth] ([shift={(-1,0.6)}]parameter_derivative.west) -- ([shift={(-4,0.6)}]parameter_derivative.west) node[pos=0.9,above]{$\bm{\eta_i}$};

        \filldraw[black] ([shift={(-1,0.6)}]parameter_derivative.west) circle (1.5pt); 

        \filldraw[black] ([shift={(0.5,0)}]parameter_derivative.east) circle (1.5pt); 

        \filldraw[black] ([shift={(0.5,0)}]integrator_1.east) circle (1.5pt); 

    \end{tikzpicture}
    }
    \caption{Planning algorithm diagram. The algorithm is run locally on each robot at a fixed rate.}
    \label{fig:block_diagram}
\end{figure*}
\section{Algorithm}\label{sec:algorithm}
The structure of the algorithm can be seen in \cref{fig:block_diagram}. The algorithm runs locally on each robot at a fixed rate, and the robots exchange information over a communication network. The algorithm relies on the parameters $\lambda_i$, $\mu_i$ and $K_{i}$ being chosen by the operator. In \cref{sec:simulation} the effect of these parameters can be seen, which can serve as a basis for hand tuning. The algorithm consists of two parts, a parameter space part and a velocity space part.
\subsection*{Parameter Space}
The parameter space part takes as input the desired velocity of the robot $\bm{v_{des,i}}$, the parameter vectors of the communication neighbours $\bm{\eta_{\mathcal{N}_i}}$, and the two multipliers $\lambda_i$ and $\mu_i$. It maps $\bm{v_{des,i}}$ into the parameter space and performs the consensus and constraint satisfaction step. From this, it generates a parameter and multiplier derivative, which it uses to update the parameters and multiplier using numerical integration, e.g. Euler integration. It then outputs the current parameters and parameter derivatives to the feedback part, and transmits the current parameters to its communication neighbours.
\subsection*{Velocity Space}
The velocity space part takes as input the parameters, parameter derivatives and the feedback gain $K_{i}$. It maps the parameter derivative back into the velocity space and performs a disturbance rejection step ensuring that the robot tracks the desired position in the formation. From this, it generates a velocity reference which it then outputs to the local kinematic controller on the robot.
\section{Simulation Results}\label{sec:simulation}
\begin{figure}[t!]
    \centering
    \scalebox{.55}{
    \includegraphics{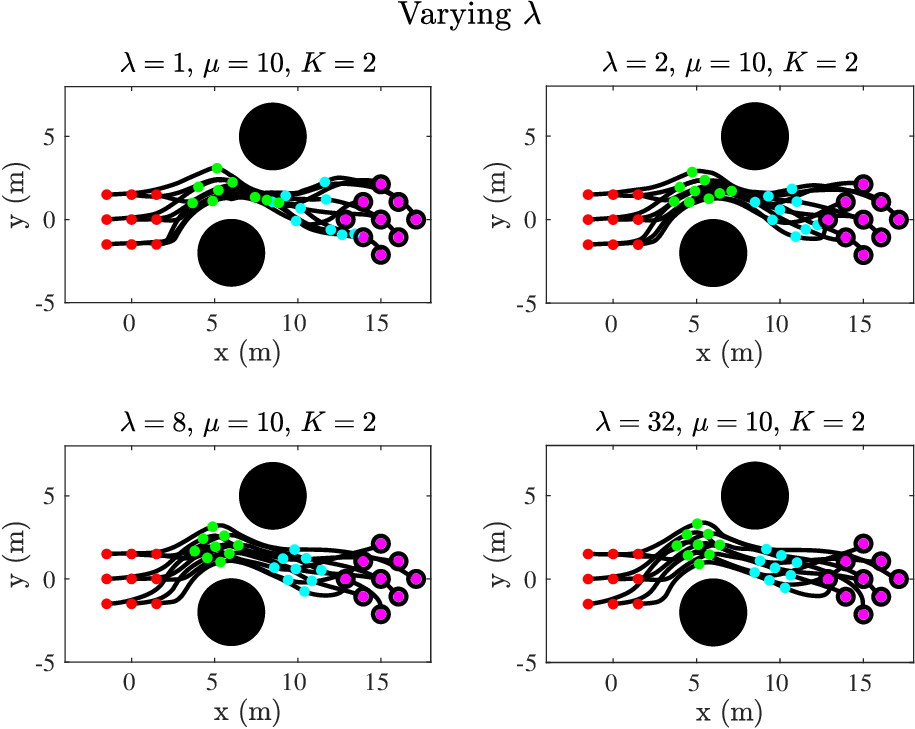}
    }
    \caption{Varying parameter $\lambda$ at four different times: \textcolor{red}{$t=0$} (red), \textcolor{green}{$t=3$} (green), \textcolor{cyan}{$t=5$} (cyan), \textcolor{magenta}{$t=8$} (magenta).}
    \label{fig:varying_lambda}
\end{figure}
To test the formation planning algorithm, a local planner on each drone must be provided. In this paper, an artificial potential field planner is used, where the obstacle closest to each robot produces a repulsive force, the position of the robot in the desired formation produces an attractive force, and since the robots themselves can be considered dynamic obstacles, the robot closest to each robot also produces a repulsive force. The desired velocity is set to the sum of these three forces, \cite{Khatib1985Real-timeRobots}. To demonstrate the effect of the parameters $\lambda$, $\mu$ and $K$, the formation planner is tested in simulation where: the base configuration is a unit grid configuration with $\bm{c_i} \in (-1,0,1)\times(-1,0,1)$; the robots start at the configuration with transformation parameters $\bm{\eta} = (0,1,1,0,0)$; the goal configuration is the configuration with transformation parameters $\bm{\eta} = (5/4\pi,1.5,1.5,15,0)$ and there are two circular obstacles with radius $2$ $m$ and centre at $(6,-2)$ $m$ and $(8.5,5)$ $m$.
The formation planner is tested in three different settings where the parameters are varied, in turn, to demonstrate their effect on the outcome of the planner. The parameters are kept identical across all robots. The remaining simulation parameters can be seen in \cref{table:sim_param}. The simulation is run in MATLAB R2022b on a Lenovo Thinkpad L15 with an AMD Ryzen 7 pro 5850u CPU with a frequency of maximum $4505$ MHz. The algorithm uses a sampling time of $1$ ms.
\subsection*{Test 1: Varying $\lambda$}
In the first test, the formation planner is simulated with $\lambda=\{1,2,8,32\}$, see \cref{fig:varying_lambda}. The magnitude of $\lambda$ has a great effect on the robots and how much they keep in formation. It is evident that as $\lambda$ increases, the robots keep the formation tighter.
\subsection*{Test 2: Varying $\mu$} 
\begin{figure}[t!]
    \centering
    \scalebox{.55}{
    \includegraphics{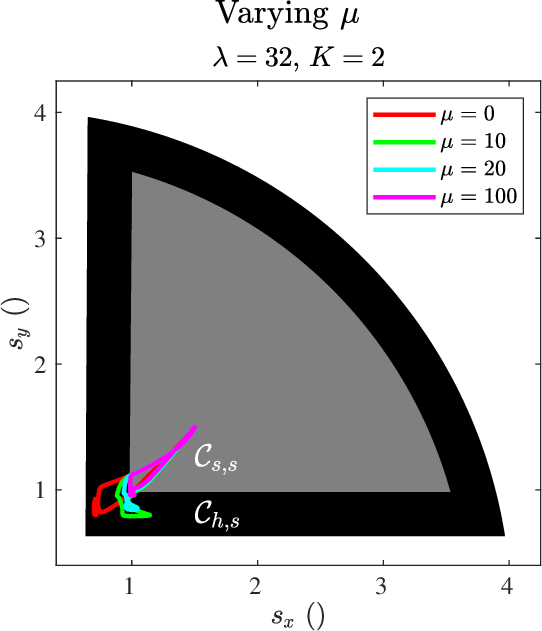}
    }
    \caption{Scaling parameter with varying parameter $\mu$.}
    \label{fig:varying_mu}
\end{figure}

In the second test, the formation planner is simulated with $\mu=\{0,10,20,100\}$, see \cref{fig:varying_mu}. The magnitude of $\mu$ has a great effect on the scaling parameters. As $\mu$ increases, the scaling parameter remains closer to the soft constraint set $\mathcal{C}_{s,s}$.
\subsection*{Test 3: Varying $K$}
\begin{figure}[t!]
    \centering
    \scalebox{.55}{
    \includegraphics{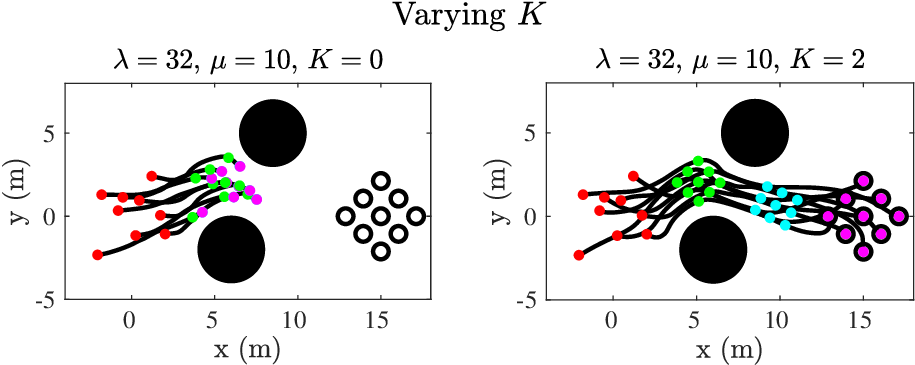}
    }
    \caption{Varying parameter $K$ at four different times: \textcolor{red}{$t=0$} (red), \textcolor{green}{$t=3$} (green), \textcolor{cyan}{$t=5$} (cyan), \textcolor{magenta}{$t=8$} (magenta).}
    \label{fig:varying_kp}
\end{figure}

In the third test, the formation planner is simulated with $K=\{0,2\}$, see \cref{fig:varying_kp}. Furthermore, in this test, the initial positions of the robots are perturbed with normal distributed random noise $\bm{\omega} \sim \mathcal{N}(\bm{0},0.5\mathbf{I_{2\times2}})$. When $K=0$, the robots cannot cancel the effect of the perturbation and return to formation, resulting in the robots being unable to reach the goal configuration. However, when $K=2$, the robots are driven towards their current desired formation and the effect of the random perturbation is cancelled, resulting in the robots terminating at the goal configuration.
\subsection*{Run-Time Efficiency}
To evaluate the applicability of the formation planning algorithm, the run-time efficiency of the simulation, for each of the steps of the algorithm, is calculated, which depends on both the hardware and the implementation of the simulation. As can be seen in \cref{tab:run-time}, the complete algorithm has a mean run time of less than $23$ $\mu$s and a maximum run time of $142$ $\mu$s, meaning that the algorithm can run at a rate of more than $7$ kHz, making it highly likely that it can be deployed in real time. Furthermore, since the simulation is run in MATLAB, it is expected that it can be greatly improved by implementing it in a more efficient programming language, such as C/C++.
\begin{table}[t!]
    \centering
    \begin{adjustbox}{max width=\linewidth}
    \begin{tabular}{c|c|c|c|c|c}
        \textbf{step} & \textbf{min} & \textbf{median} & \textbf{mean} & \textbf{max} & \textbf{variance} \\
        \hline
        \hline
        tracking & $5e-6$ & $7e-6$ & $7.1e-6$ & $4.4e-5$ & $2.4e-12$ \\
        \hline
        consensus & $3e-6$ & $4e-6$ & $4.03e-6$ & $2.6e-5$ & $7.03e-13$ \\
        \hline
        soft constraint & $1e-6$ & $2e-6$ & $1.76e-6$ & $1.7e-5$ & $3.96e-13$ \\
        \hline
        hard constraint & $4e-6$ & $5e-6$ & $5.26e-6$ & $3.1e-5$ & $1.12e-12$ \\
        \hline
        recovering velocity & $4e-6$ & $4e-6$ & $4.43e-6$ & $2.4e-5$ & $8.35e-13$ \\
        \hline
        complete method & $1.7e-5$ & $2.2e-5$ & $2.26e-5$ & $1.42e-04$ & $5.45e-12$
    \end{tabular}
    \end{adjustbox}
    \caption{Run-time of the formation planning steps and complete method in seconds.}
    \label{tab:run-time}
\end{table}

\section{Conclusions}
This paper presented a distributed motion planning algorithm for rigid formations. The planner takes as input the local desired velocities of the robots and calculates the alterations to the velocities that ensure that they keep formation. The algorithm is able to handle hard constraints, and its efficacy and run-time efficiency have been evaluated. The efficacy of the algorithm has been shown through simulation results proving it to be a promising approach for deployment on robots with limited computational ability. 

\printbibliography

\appendix

\subsection{Simulation Parameters}

\begin{table}[h]
\centering
    \begin{tabular}{c | c | c | c} 
        \textbf{Parameter} & \textbf{Value} & \textbf{Unit} & \textbf{Description} \\
        \hline
        \hline
        $t_s$ & $1\e-3$ & s & sampling time \\
        $t_{final}$ & $9$ & s & simulation time \\
        $\varepsilon$ & $0.75$ & - & constraint parameter \\
        $r_{max}$ & $2.5$ & - & constraint parameter \\
        $k_{att}$ & $5$ & m/s & attractor velocity \\
        $\rho$ & $0.1$ & m &  attractor switch distance \\
        $k_{rep}$ & $5$ & m/s & repulsor strength \\
        $\xi$ & $0.25$ & m & obstacle clearance \\
        $\nu$ & $1.5$ & m & repulsor distance 
    \end{tabular}
    \caption{Simulation parameters.}
    \label{table:sim_param}
\end{table}

\end{document}